\documentclass[runningheads]{llncs}
\usepackage[T1]{fontenc}
\usepackage{graphicx}
\usepackage{comment}
\usepackage[a4paper]{geometry}
\usepackage{multirow}
\usepackage{amsmath}
\usepackage{float}
\usepackage[caption=false]{subfig}
\usepackage{capt-of}
\usepackage[hidelinks]{hyperref}
\begin{document}

\title{ChurnNet:  A Optimized Modern AI for Churn Prediction}
\titlerunning{Traditional ML Methods vs UniTS for Retail Churn Prediction}
\author{Syed Saad Saif\inst{1,2}\orcidID{0009-0003-4267-7060} 
\and
Giulio Maggiore\inst{3}\orcidID{0000-0002-4657-0362}
\and
Paolo Russo\inst{4}\orcidID{0000-0002-1886-3491}
\and
Damiano Distante\inst{3}\orcidID{0000-0002-8467-535X}
}
\authorrunning{S. Saad Saif et al.}
%
%

\institute{Department of Computer, Control, and Management Engineering, Sapienza University of Rome, Rome, Italy 
\and
Department of Law and Economics, UnitelmaSapienza University of Rome, Rome, Italy
\and
R\&D Center, Token Financial Technologies, Istanbul, Turkey
\and
Department of Civil, Computer Science and Aeronautical Technologies Engineering, \protect\\Roma Tre University, Rome, Italy
}

\maketitle              
\begin{abstract}
Increased competition and the growing similarity of products and services offered by retailers have lowered the barriers for customers to switch to competitors. Accurate churn prediction can be a valuable tool for driving effective personalized marketing campaigns and helping to reduce customer attrition.
This study evaluates the performance of traditional machine learning techniques, namely, Random Forests, XGBoost, and Support Vector Machines, and compares them with the Unified Multi-Task Time Series Model for churn prediction, a binary time-series classification task. Despite the strong capacity of the latter to model complex temporal dynamics and inter-variable relationships, our results indicate that for churn prediction, conventional methods can still outperform it in terms of predictive performance, data efficiency, and computational resource requirements for training and deployment. These findings are consistent across multiple datasets and various churn labeling techniques.
\keywords{Artificial intelligence \and customer churn prediction \and retail sector \and non-contractual environment \and Unified Multi-Task Time Series Model \and UniTS \and Random Forest \and XGBoost \and Support Vector Machine \and SVM}
\end{abstract} 
\section{Introduction}

In recent years, the retail industry has undergone a profound transformation. The adoption of online as well as omni-channel sales has allowed customers to have convenient access to multiple retailers through digital platforms, unlike traditional retail,  which relied primarily on physical touch-points and local customer bases. This evolution has intensified competition and dramatically lowered switching barriers \cite{Reinartz2003TheIO,basker2012supersize} making customer retention a strategic priority for retailers of all sizes.

Customer churn, the process by which consumers reduce or cease their purchasing activity, has emerged as a critical challenge in this context. In non-contractual settings, such as e-commerce and grocery retail, churn does not occur through formal cancellations but through behavioral changes, such as decreased purchase frequency or volume \cite{S37-ID826-Buckinx}. Detecting these gradual forms of defection is particularly challenging yet crucial, as subtle declines in customer engagement often precede full attrition \cite{Singh2021,Gupta2021}.

At the same time, the increasing availability of customer data, particularly through loyalty programs, digital receipts, browsing histories, and CRM systems has enabled more sophisticated approaches to customer analytics \cite{Zhang2022,Hagberg2016}. This abundance of data has catalyzed the application of machine learning (ML) and artificial intelligence (AI) methods for churn prediction, which in turn can help businesses to identify early warning signals and implement targeted retention strategies \cite{Gupta2021,Lee2020-SpatialApproach}.

 AI‑based churn prediction has been extensively explored in \textit{contractual industries}, such as telecommunications, banking, and insurance, with several recent surveys reviewing both traditional ML and deep learning approaches (e.g., \cite{Geiler2022,Manzoor2024,Barsotti2024}). However, the \textit{non‑contractual retail sector} remains relatively under investigated. Differently from contractual contexts where churn is defined by discrete events such as service cancellations, in non-contractual retail the absence of explicit termination complicates both model design and the operationalization of churn labels \cite{Chen2019,Rodríguez2023}. 

In order to develop a churn prediction system, the question regarding the most suitable machine learning approach naturally arises. On one hand, recent techniques, such as UniTS \cite{Gao2024units}, TimesNet \cite{wu2023timesnet}, and FlowFormer \cite{wu2022flowformer}, show impressive performance for binary time-series classification tasks on datasets of varying domains and levels of difficulty. To achieve this performance, these models adopt a deep and complex architecture and, therefore, are computationally expensive and prone to overfitting \cite{benkendorf2020effects}. To mitigate this issue, they are required to be trained on large datasets. 
In contrast, conventional techniques have also been successfully applied to achieve good performance in churn prediction. These models, thanks to their relatively simple architectures, are more computationally and data-efficient. 
In this paper, we examine whether the comparatively simple nature of churn prediction as a time-series binary classification task necessitates the use of deep, modern, but complex architectures, or if traditional, simpler models are sufficient to deliver strong predictive performance.




For this analysis, we compared traditional machine learning models, namely, Random Forests, XGBoost and Support Vector Machines (SVM), with the Unified Multi-Task Time Series Model (UniTS) \cite{Gao2024units}, a transformer-based model specifically designed to handle temporal data. We chose the UniTS model to represent modern deep learning architectures, as it delivers top performance on the time-series binary classification task \cite{Gao2024units}. The study spans multiple real-world retail datasets and employs a family of churn labeling definitions based on relative drops in average weekly spending over time windows. Our goal is to assess not only the raw predictive performance (precision, recall, F1-score) but also to analyze it across datasets with varying characteristics, including size and levels of imbalance. Such an analysis is important when considering systems intended for real-world deployment. In fact, the results of this study will inform the development of a commercial churn prediction system by Infocube\footnote{\url{https://www.infocube.it}}, a digital marketing solutions provider, which supplied us with two of the datasets included in the evaluation suite of this research.

The remainder of this paper is structured as follows. Section ~\ref{sec:relatedwork} reviews relevant studies on AI-based churn prediction in e-commerce and retail, with a focus on applied models, recent innovations, and the use of advanced AI techniques. Section~\ref{sec:background} provides an overview of customer churn predictive models used in this study, discussing their theoretical suitability for the churn prediction task. Section~\ref{sec:methodology} details our methodology, including the preparation of the datasets and the churn labeling strategy. Section~\ref{sec:implementation} describes the implementation of the models. Section~\ref{sec:results} presents the empirical results,  offering a discussion on the performance and interpretation of the results, as well as comments on the implications of each method in practical churn prediction scenarios. Section~\ref{sec:threats} discusses the limitations and threats to the validity of this study. Finally, Section~\ref{sec:conclusion} concludes the paper and outlines potential directions for future research.

\section{Related Works} 
\label{sec:relatedwork}

Although research on churn prediction has predominantly focused on contractual environments, a number of studies have also examined this task in non-contractual settings. These studies employ a wide range of AI techniques. While the literature is largely centered on traditional methods, such as Decision Trees (DT), Random Forests (RF), XGBoost (XGB), SVM, K-Nearest Neighbors (KNN), and Logistic Regression (LR), more recent approaches, including Artificial Neural Networks (ANN), Convolutional Neural Networks (CNN), and Long Short-Term Memory (LSTM) networks, have been applied with varying degrees of success \cite{Shahabikargar10072025,Manzoor2024,CiucciColli2025}.

Rahib et al. \cite{S1-ID3-Rahib} applied several traditional machine learning methods such as SVM, RF and DT. Sunarya et al. \cite{S11-D105-Sunarya} implemented LR and RF. Agbemadon et al. \cite{S19-ID363-Agbemadon} applied both conventional AI techniques, like LR and XGB, as well as modern approaches such as ANN and LSTM.
Boukrouh et al. \cite{S7-ID79-Boukrouh} conducted a comprehensive analysis of a broad suite of AI techniques that included  DT, RF, SVM, LR, NB, KNN, and ANN. In addition to evaluating predictive performance this study also and utilized explainable artificial intelligence (XAI) methods such as Shapley Additive Explanations (SHAP) and Local Interpretable Model-agnostic Explanations (LIME) to develop interpretations of these models.

Beyond the application of standard techniques, some studies have focused on methodological innovations aimed at enhancing AI model performance. Gan et al. \cite{S17-ID344-Gan} integrated social behavior and influence dynamics into the XGBoost framework to capture the impact of customers on their social networks. Pondel et al. \cite{S23-ID396-Pondel} focused on e-commerce data, incorporating non-time-series departmental information exprimenting with variations of ANN and Recurrent Neural Networks (RNN) to identify the best-performing option.
Miguéis et al. \cite{S35-ID783-Miguéis} explored the implementation of Multivariate Adaptive Regression Splines (MARS) and benchmarked its performance against LR. Notably, Seema et al. \cite{S10-ID98-Seema} introduced a novel approach involving a fading channel patch-based heat map for training CNN, aiming to improve deep learning model effectiveness in the churn prediction context.

While previous studies have explored a range of AI techniques for churn prediction, most focus on either traditional or complex models in isolation. Our work stands out by comparing traditional machine learning models, such as Decision Trees, Random Forests, and XGBoost, with UniTS, a transformer-based model specifically designed for time-series data. This unique comparison highlights the effectiveness of simpler models versus those tailored for temporal tasks, filling a gap in the existing literature.
\section{Background} 
\label{sec:background}

\subsection{Definition of Churn}

In the retail sector, customer churn refers to the phenomenon in which a previously active customer ceases or substantially reduces engagement in purchasing behavior over a defined period. Unlike contractual settings, where churn is explicitly observed through contract terminations, churn in retail is typically inferred from behavioral inactivity, which makes its detection more challenging. 

Labeling customers as churners in retail typically involves defining a period of inactivity or significant drop in engagement, known as the churn window or churn horizon, beyond which a customer is considered lost. Common approaches include fixed thresholds, such as no purchases within 3, 6, or 12 months \cite{Gangadhar2023,Baghla2022}, or dynamic thresholds based on individual purchase patterns, such as multiples of the customer's average inter-purchase time or spend \cite{Rachid2018,Pondel2021,9670561}. Some studies also apply customer segmentation and clustering approaches, probability-based scoring, or survival analysis to estimate churn probabilities \cite{Lee2024,Coolwijk2024}. 

Accurate churn labeling is critical, as it influences both model training and the business interpretation of churn risk. However, definitions vary widely across studies and industries.

\subsection{Utilized AI Techniques}

As highlighted by several literature reviews \cite{Shahabikargar10072025,Manzoor2024} and our previous study \cite{CiucciColli2025}, the traditional machine learning models selected in this work are among the most widely used and well-established AI techniques in the churn prediction literature. 
In contrast, UniTS represents a modern deep learning model specifically designed for time-series data that has been successfully applied to a variety of domains \cite{Gao2024units}.

Below, we briefly describe the AI techniques used in this study, including both traditional machine learning methods and UniTS, highlighting their key characteristics and relevance to churn prediction.

\subsubsection{Random Forests}

RF, introduced by Breiman in 2001, addresses overfitting in decision trees by aggregating predictions from multiple trees built on bootstrapped samples and random subsets of features \cite{breiman2001random}. Each tree in the forest votes on the outcome, and the majority vote determines the final prediction. This ensemble approach enhances stability and accuracy compared to a single tree. Random Forests are particularly effective in handling high-dimensional, nonlinear data and offer valuable insights into feature importance. In churn prediction, Random Forests can capture complex patterns in customer behavior; however, it may be less effective at modeling time dependencies compared to sequential models.

\subsubsection{XGBoost}

XGBoost, released by Chen in 2016, is an efficient gradient boosting framework designed to improve speed and performance through parallelization and regularization \cite{chen2016xgboost}. It builds additive decision trees sequentially, where each new tree corrects the errors made by the previous ensemble using gradient information from a loss function. XGBoost integrates techniques such as shrinkage, column subsampling, and tree pruning to prevent overfitting and enhance performance. It achieves high accuracy on tabular data, handles missing values, and manages class imbalance effectively. However, to extract its best performance, careful fine-tuning is required. In churn prediction, XGBoost has proven to be particularly effective in capturing complex patterns in customer behavior and handling imbalanced datasets.

\subsubsection{Support Vector Machines}

SVM, introduced in the early 1990s, aim to identify optimal separating hyperplanes for classification tasks, using kernel functions to model nonlinear relationships \cite{cortes1995support}. The core idea is to find the hyperplane that maximizes the margin between different classes, while support vectors (data points closest to the boundary) define this separation. Kernel tricks transform data into higher-dimensional spaces to make it linearly separable when needed. SVMs are powerful in high-dimensional spaces but may encounter scalability issues with large datasets and are particularly sensitive to class imbalance, a common challenge in churn prediction.

\subsubsection{UniTS}

UniTS, introduced in 2024 by Gao et al. \cite{Gao2024units}, is a unified multi-task time-series model designed to handle forecasting, classification, anomaly detection, and imputation within a single architecture. Inspired by foundation models in Natural Language Processing (NLP) and computer vision, it incorporates task-tokenization, attention mechanisms, and dynamic linear operators to enable parameter sharing and adaptation across tasks. UniTS processes input sequences through transformers equipped with specialized tokens that signal different tasks, allowing the same network to adapt flexibly to diverse objectives. UniTS shows strong performance across diverse datasets but requires significant computational resources and large-scale pre-training. Its capacity to capture complex temporal patterns makes it theoretically well-suited for modeling customer behavior over time, such as in churn prediction.
\section{Methodology} 
\label{sec:methodology}

Our methodology follows a structured pipeline designed to prepare transactional data, generate churn labels, and evaluate multiple machine learning models. An overview of the entire training and evaluation process is presented in Figure~\ref{fig:training_evaluation_pipeline}.

The process begins with dataset pre-processing and filtering, where raw transactional data is cleaned and transformed into time series for each customer. Next, customers are labeled based on different churn definitions. For model training, temporal windows are selected. Various balancing and sub-sampling strategies are subsequently applied. This produces multiple representative training datasets, which are then used to train machine learning models and evaluate their performance. The following subsections provide a detailed explanation each step. 

\subsection{Dataset Pre-processing}
The goal of this step is to clean the data, remove anomalies, and establish a standardized and consistent starting point for analysis. During pre-processing, transactional data is aggregated by customer ID and filtered to exclude non-positive values. Additionally, transaction amounts are aggregated on a weekly basis to ensure temporal consistency across the dataset. The dataset is also converted into time series data by zero-padding the weeks where no transactions occurred for a customer.

\subsection{Dataset Filtering}
To ensure the reliability of predictions, we focus on customers with a sufficient transactional history. Customers with limited activity provide insufficient behavioral patterns for meaningful analysis and are typically of lower commercial interest. Therefore, in our study, we exclude customers with fewer than 20 transactional weeks i.e., weeks during which the customer made at least one transaction.

\newsavebox{\LeftBox}
\begin{lrbox}{\LeftBox}
\begin{minipage}[t]{0.58\textwidth}
    \centering
    \vspace{5mm}
    \includegraphics[width=\linewidth]{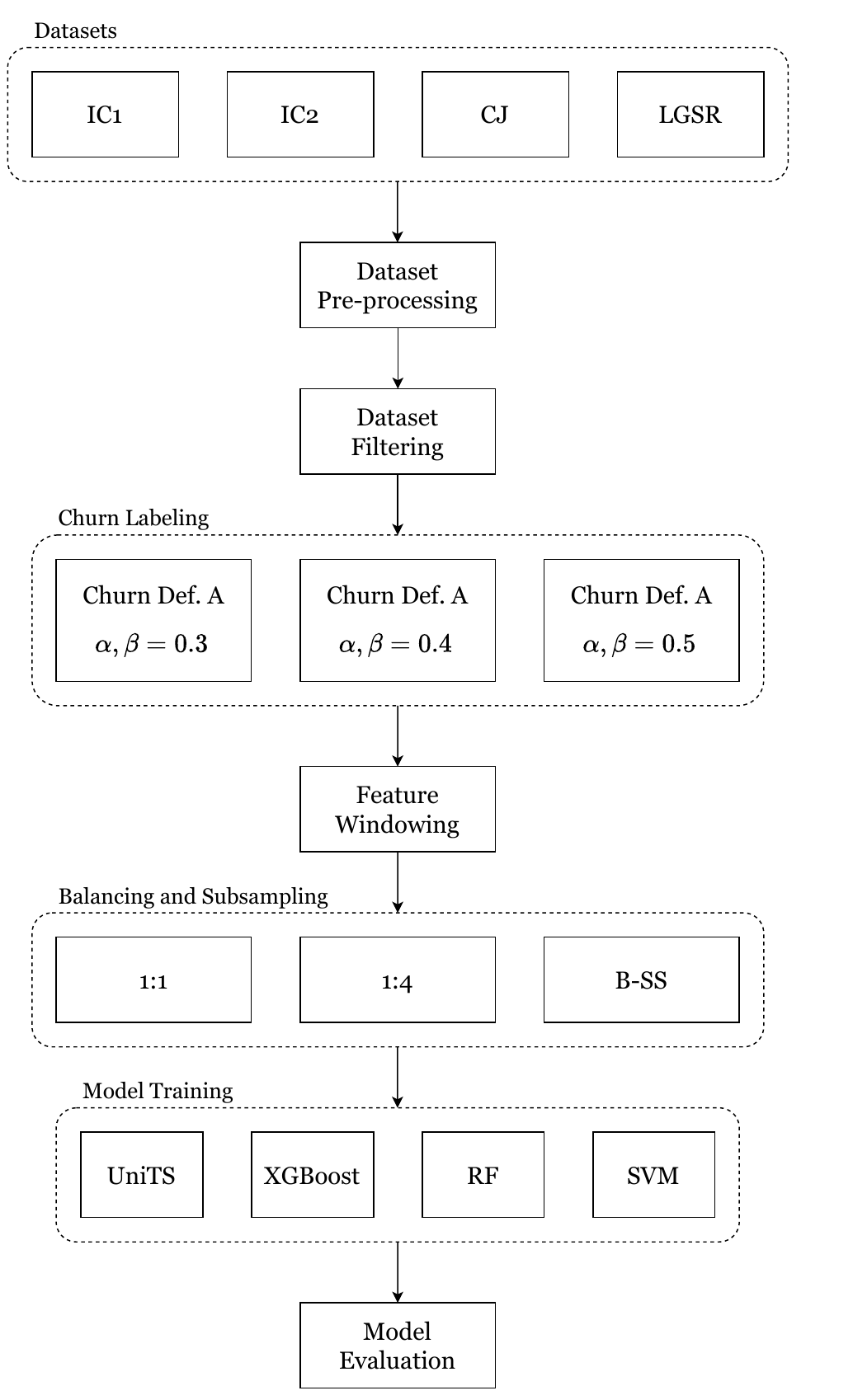}
    \vspace{5mm}
    \captionof{figure}{Data processing, training and evaluation pipeline of the proposed methodology.}
    \label{fig:training_evaluation_pipeline}
\end{minipage}
\end{lrbox}

\noindent
\usebox{\LeftBox}
\hfill
\begin{minipage}[t]{0.40\textwidth}
\centering
\vbox to \ht\LeftBox{%
\begin{minipage}{\linewidth}
\centering
\captionof{table}{Churn Labeling Definitions.}
\vspace{6pt}
\label{tab:churn_definitions}
\begin{tabular}{|c|c|c|}
\hline
\textbf{Churn Definition} & \textbf{\(\alpha\)} & \textbf{\(\beta\)} \\
\hline
A & 0.3 & 0.3 \\
B & 0.4 & 0.4 \\
C & 0.5 & 0.5 \\
\hline
\end{tabular}%
\end{minipage}
\par\vspace{4mm}
\begin{minipage}{\linewidth}
\centering

\begin{minipage}{\linewidth}
\centering
\includegraphics[width=\linewidth]{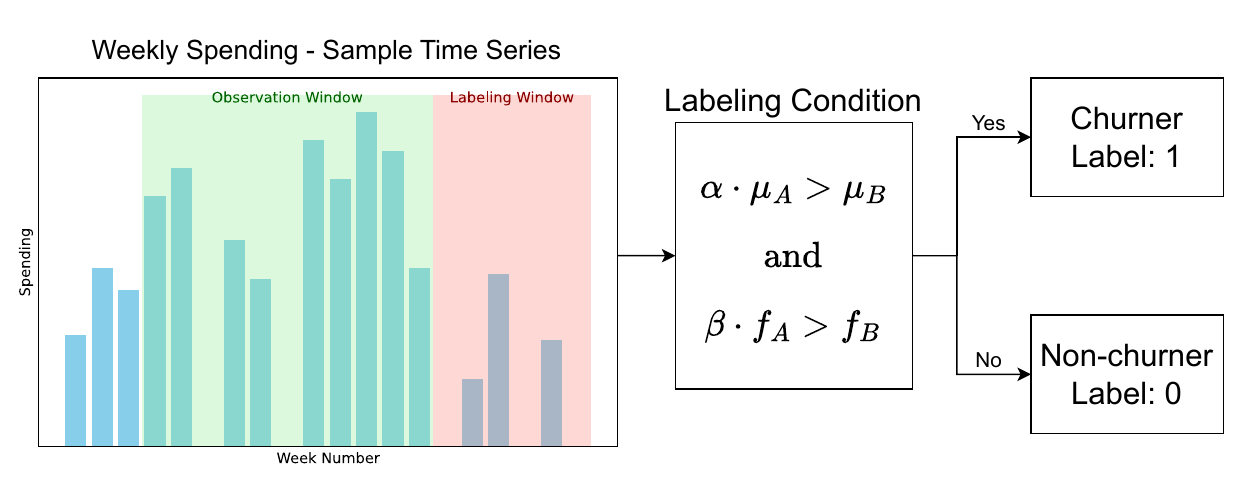}
\captionof{figure}{Churn labeling strategy.}
\label{fig:churn_labeling}
\end{minipage}

\par\vspace{4mm}

\includegraphics[width=\linewidth]{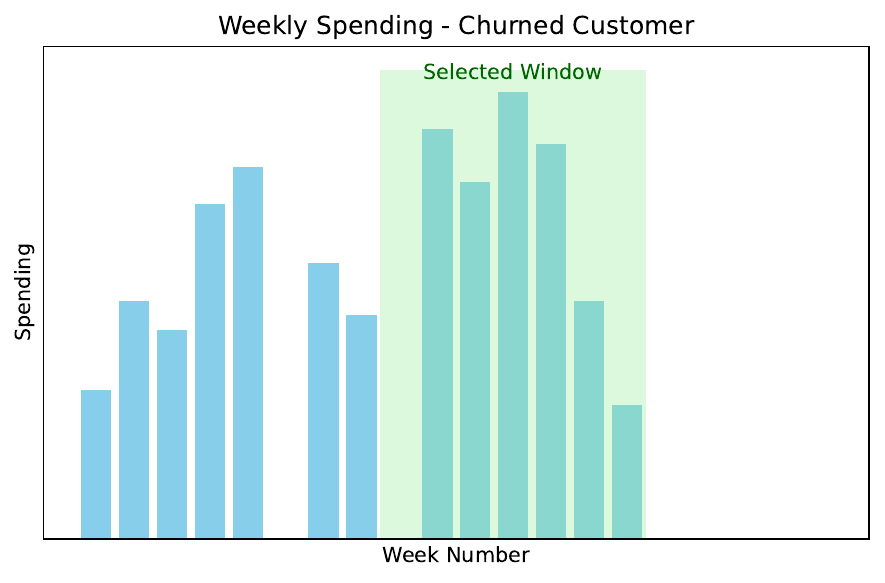}\\
\footnotesize(a) Churn customer.
\label{fig:windowing_churner}

\vspace{2mm}

\includegraphics[width=\linewidth]{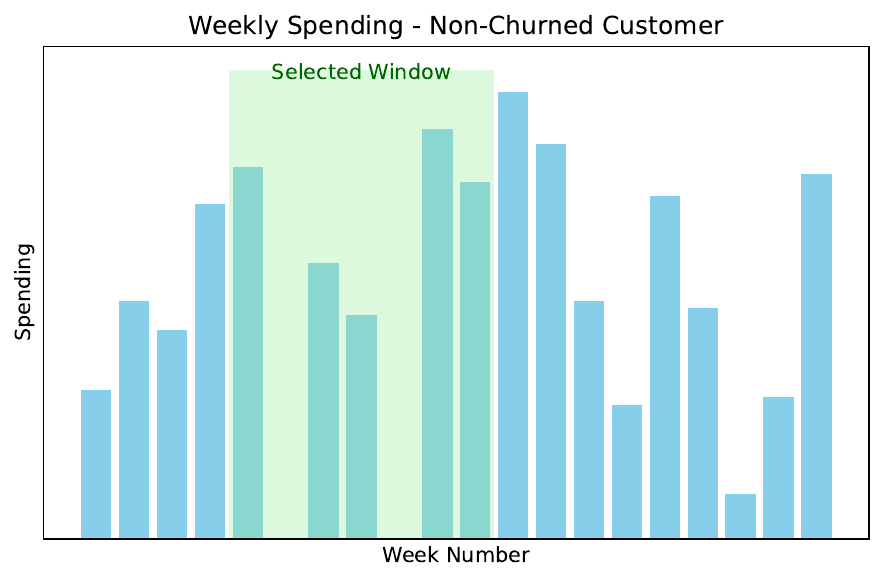}\\
\footnotesize(b) Non-churn customer.
\label{fig:windowing_nonchurner}

\captionof{figure}{Feature windowing strategies.}
\label{fig:feature_windowing}
\end{minipage}

}
\end{minipage}

\subsection{Churn Labeling}
\label{subsec:churn_labeling}
As discussed in the background section \ref{sec:background}, churn labeling in non-contractual settings is a non-trivial task, with various approaches available depending on the use case and data characteristics. In this work, we adopt a dynamic churn definition, formulated in collaboration with Infocube and tailored to the needs of their retail clients. This approach aims to provide a more accurate and individualized churn label by accounting for each customer's historical shopping behavior.

We define two key timeframes: an \textit{observation window} of length \( A = 40 \) weeks and a subsequent \textit{labeling window} of length \( B = 20 \) weeks. The two windows are contiguous and are aligned with the end date of their respective dataset. The observation window is used to calculate personalized thresholds which are then applied to the labeling window to determine each customer’s churn status.

The labeling logic is based on two behavioral parameters, which are elaborated below:
\begin{itemize}
    \item \textbf{Monetary value}: The average weekly spending in the observation window is denoted by \( \mu_A \), and in the labeling window by \( \mu_B \).
    \item \textbf{Frequency of activity}: The average number of transactions per week in the observation window is denoted by \( f_A \), and by \( f_B \) in the labeling window.
\end{itemize}

A customer is labeled as a \textit{churner} (1) if both of the following conditions are met:

\begin{equation}
    \alpha \cdot \mu_A > \mu_B \quad \text{and} \quad \beta \cdot f_A > f_B
    \label{eq:churn_labeling}
\end{equation}

Otherwise, the customer is labeled as a \textit{non-churner} (0). Here, \( \alpha \) and \( \beta \) are configurable threshold parameters, which control the sensitivity of the churn detection logic. By evaluating changes relative to each customer's historical behavior, this method reduces noise and provides a more robust and personalized definition of churn.

A visual illustration of this labeling method, showing the positioning of the observation and labeling windows, is provided in Figure~\ref{fig:churn_labeling}.

We employed three different churn labeling definitions by adjusting the sensitivity parameters \(\alpha\) and \(\beta\), as summarized in Table~\ref{tab:churn_definitions}. These variants enable us to evaluate the robustness of our models under different churn detection thresholds.


\subsection{Feature Windowing}
The feature window refers to the time period over which input features (\( X \)) are passed to our models for training. In our study, this window is set to a fixed length of 20 weeks. The selection of the feature window depends on the churn label assigned to the customer:

\begin{itemize}
    \item \textbf{For churners}, the feature window ends immediately before the customer's final period of sustained inactivity.
    \item \textbf{For non-churners}, a random valid segment of 20 weeks is selected from the customer's available transaction history.
\end{itemize}

This strategy is motivated by the assumption that, for churners, churn-related behavioral patterns are likely to be more pronounced in the period leading up to the churn event. Therefore, capturing this window enables the model to learn relevant dynamics associated with customer disengagement. In contrast, for non-churners, there is no a priori knowledge about when, or if, such patterns might occur, so a random 20-week segment is selected to provide a representative sample of normal activity.

An illustrative example of the feature windowing process for both churners and non-churners is provided in Figure~\ref{fig:feature_windowing}.

\subsection{Balancing and Subsampling}
\label{subsec:balance_and_subsample}

To assess the impact of dataset size and class imbalance on model performance, we employed the following dataset balancing and sub-sampling strategies:

\begin{itemize}
    \item \textbf{Balanced Dataset (1:1)}:  
    These datasets include all churners remaining after filtering. An equal number of non-churners are then randomly sampled to create a balanced dataset. This setting is designed to neutralize the impact of class imbalance on model performance. Although this reduces the total amount of data used for training, it is less of a concern for large-scale datasets.

    \item \textbf{Limited Unbalanced Dataset (1:4)}:  
    In this setting, all filtered churners are retained, and four times as many non-churners are randomly sampled. This controlled imbalance allows us to evaluate model performance under moderately imbalanced conditions while still utilizing more training data. It provides insight into how well models handle class imbalance in a realistic yet constrained scenario.

    \item \textbf{Balanced Subsampled Dataset (B-SS)}:  
    To investigate the data efficiency of different models, especially on smaller datasets, we created downsampled versions of the larger datasets. For \textit{single-star (*)} datasets, we randomly select 2,500 customers after filtering; all churners are included, and an equal number of non-churners are sampled. For \textit{double-star (**)} datasets, the same process is applied with 5,000 customers. This setting allows us to isolate the impact of dataset size on performance while maintaining balance.
\end{itemize}

\begin{table}[t]
\centering
\caption{Dataset Characteristics.}
\label{tab:data_desc_2}
\resizebox{0.54\textwidth}{!}{
\begin{tabular}{ccc|ccc|c|rrrr|}
\cline{4-11}
                                                                                    &                                                     &                & \multicolumn{3}{c|}{\multirow{2}{*}{\textbf{\begin{tabular}[c]{@{}c@{}}Normalized\\ Tenure\end{tabular}}}} & \multirow{2}{*}{\textbf{Ratio}} & \multicolumn{4}{c|}{\textbf{Number of Customer}}                                                                                              \\ \cline{8-11} 
                                                                                    &                                                     &                & \multicolumn{3}{c|}{}                                                                                      &                                 & \multicolumn{2}{c|}{\textbf{Before Balancing}}                        & \multicolumn{2}{c|}{\textbf{After Balancing}}                         \\ \hline
\multicolumn{1}{|c|}{\textbf{\begin{tabular}[c]{@{}c@{}}Churn\\ Def.\end{tabular}}} & \multicolumn{1}{c|}{\textbf{Cat.}}                  & \textbf{Data}  & \multicolumn{1}{c|}{\textbf{Total}}        & \multicolumn{1}{c|}{\textbf{CH}}        & \textbf{N-CH}       & \textbf{CH}                     & \multicolumn{1}{c|}{\textbf{CH}} & \multicolumn{1}{c|}{\textbf{N-CH}} & \multicolumn{1}{c|}{\textbf{CH}} & \multicolumn{1}{c|}{\textbf{N-CH}} \\ \hline
\multicolumn{1}{|c|}{\multirow{12}{*}{\textbf{A}}}                                  & \multicolumn{1}{c|}{\multirow{4}{*}{\textbf{1:1}}}  & \textbf{IC1}   & \multicolumn{1}{c|}{0.92}                  & \multicolumn{1}{c|}{0.69}               & 0.94                & 0.08                            & \multicolumn{1}{r|}{12,425}      & \multicolumn{1}{r|}{143,454}       & \multicolumn{1}{r|}{12,425}      & 12,425                             \\ \cline{3-11} 
\multicolumn{1}{|c|}{}                                                              & \multicolumn{1}{c|}{}                               & \textbf{IC2}   & \multicolumn{1}{c|}{0.89}                  & \multicolumn{1}{c|}{0.71}               & 0.90                & 0.09                            & \multicolumn{1}{r|}{35,309}      & \multicolumn{1}{r|}{360,585}       & \multicolumn{1}{r|}{35,309}      & 35,309                             \\ \cline{3-11} 
\multicolumn{1}{|c|}{}                                                              & \multicolumn{1}{c|}{}                               & \textbf{CJ}    & \multicolumn{1}{c|}{0.90}                  & \multicolumn{1}{c|}{0.75}               & 0.90                & 0.03                            & \multicolumn{1}{r|}{72}          & \multicolumn{1}{r|}{2,060}         & \multicolumn{1}{r|}{72}          & 72                                 \\ \cline{3-11} 
\multicolumn{1}{|c|}{}                                                              & \multicolumn{1}{c|}{}                               & \textbf{LGSR}  & \multicolumn{1}{c|}{0.94}                  & \multicolumn{1}{c|}{0.76}               & 0.94                & 0.01                            & \multicolumn{1}{r|}{1,500}       & \multicolumn{1}{r|}{283,880}       & \multicolumn{1}{r|}{1,500}       & 1,500                              \\ \cline{2-11} 
\multicolumn{1}{|c|}{}                                                              & \multicolumn{1}{c|}{\multirow{4}{*}{\textbf{1:4}}}  & \textbf{IC1}   & \multicolumn{1}{c|}{0.92}                  & \multicolumn{1}{c|}{0.69}               & 0.94                & 0.08                            & \multicolumn{1}{r|}{12,425}      & \multicolumn{1}{r|}{143,454}       & \multicolumn{1}{r|}{12,425}      & 49,700                             \\ \cline{3-11} 
\multicolumn{1}{|c|}{}                                                              & \multicolumn{1}{c|}{}                               & \textbf{IC2}   & \multicolumn{1}{c|}{0.89}                  & \multicolumn{1}{c|}{0.71}               & 0.90                & 0.09                            & \multicolumn{1}{r|}{35,309}      & \multicolumn{1}{r|}{360,585}       & \multicolumn{1}{r|}{35,309}      & 141,236                            \\ \cline{3-11} 
\multicolumn{1}{|c|}{}                                                              & \multicolumn{1}{c|}{}                               & \textbf{CJ}    & \multicolumn{1}{c|}{0.90}                  & \multicolumn{1}{c|}{0.75}               & 0.90                & 0.03                            & \multicolumn{1}{r|}{72}          & \multicolumn{1}{r|}{2,060}         & \multicolumn{1}{r|}{72}          & 288                                \\ \cline{3-11} 
\multicolumn{1}{|c|}{}                                                              & \multicolumn{1}{c|}{}                               & \textbf{LGSR}  & \multicolumn{1}{c|}{0.94}                  & \multicolumn{1}{c|}{0.76}               & 0.94                & 0.01                            & \multicolumn{1}{r|}{1,500}       & \multicolumn{1}{r|}{283,880}       & \multicolumn{1}{r|}{1,500}       & 6,000                              \\ \cline{2-11} 
\multicolumn{1}{|c|}{}                                                              & \multicolumn{1}{c|}{\multirow{4}{*}{\textbf{B-SS}}} & \textbf{IC1*}  & \multicolumn{1}{c|}{0.92}                  & \multicolumn{1}{c|}{0.69}               & 0.94                & 0.08                            & \multicolumn{1}{r|}{12,425}      & \multicolumn{1}{r|}{143,454}       & \multicolumn{1}{r|}{390}         & 390                                \\ \cline{3-11} 
\multicolumn{1}{|c|}{}                                                              & \multicolumn{1}{c|}{}                               & \textbf{IC2*}  & \multicolumn{1}{c|}{0.89}                  & \multicolumn{1}{c|}{0.71}               & 0.90                & 0.09                            & \multicolumn{1}{r|}{35,309}      & \multicolumn{1}{r|}{360,585}       & \multicolumn{1}{r|}{416}         & 416                                \\ \cline{3-11} 
\multicolumn{1}{|c|}{}                                                              & \multicolumn{1}{c|}{}                               & \textbf{IC1**} & \multicolumn{1}{c|}{0.92}                  & \multicolumn{1}{c|}{0.69}               & 0.94                & 0.08                            & \multicolumn{1}{r|}{12,425}      & \multicolumn{1}{r|}{143,454}       & \multicolumn{1}{r|}{200}         & 200                                \\ \cline{3-11} 
\multicolumn{1}{|c|}{}                                                              & \multicolumn{1}{c|}{}                               & \textbf{IC2**} & \multicolumn{1}{c|}{0.89}                  & \multicolumn{1}{c|}{0.71}               & 0.90                & 0.09                            & \multicolumn{1}{r|}{35,309}      & \multicolumn{1}{r|}{360,585}       & \multicolumn{1}{r|}{220}         & 220                                \\ \hline
\multicolumn{1}{|c|}{\multirow{12}{*}{\textbf{B}}}                                  & \multicolumn{1}{c|}{\multirow{4}{*}{\textbf{1:1}}}  & \textbf{IC1}   & \multicolumn{1}{c|}{0.92}                  & \multicolumn{1}{c|}{0.74}               & 0.94                & 0.11                            & \multicolumn{1}{r|}{16,808}      & \multicolumn{1}{r|}{139,071}       & \multicolumn{1}{r|}{16,808}      & 16,808                             \\ \cline{3-11} 
\multicolumn{1}{|c|}{}                                                              & \multicolumn{1}{c|}{}                               & \textbf{IC2}   & \multicolumn{1}{c|}{0.89}                  & \multicolumn{1}{c|}{0.74}               & 0.90                & 0.11                            & \multicolumn{1}{r|}{43,291}      & \multicolumn{1}{r|}{352,603}       & \multicolumn{1}{r|}{43,291}      & 43,291                             \\ \cline{3-11} 
\multicolumn{1}{|c|}{}                                                              & \multicolumn{1}{c|}{}                               & \textbf{CJ}    & \multicolumn{1}{c|}{0.90}                  & \multicolumn{1}{c|}{0.79}               & 0.90                & 0.06                            & \multicolumn{1}{r|}{124}         & \multicolumn{1}{r|}{2,008}         & \multicolumn{1}{r|}{124}         & 124                                \\ \cline{3-11} 
\multicolumn{1}{|c|}{}                                                              & \multicolumn{1}{c|}{}                               & \textbf{LGSR}  & \multicolumn{1}{c|}{0.94}                  & \multicolumn{1}{c|}{0.78}               & 0.94                & 0.01                            & \multicolumn{1}{r|}{3,338}       & \multicolumn{1}{r|}{282,042}       & \multicolumn{1}{r|}{3,338}       & 3,338                              \\ \cline{2-11} 
\multicolumn{1}{|c|}{}                                                              & \multicolumn{1}{c|}{\multirow{4}{*}{\textbf{1:4}}}  & \textbf{IC1}   & \multicolumn{1}{c|}{0.92}                  & \multicolumn{1}{c|}{0.74}               & 0.94                & 0.11                            & \multicolumn{1}{r|}{16,808}      & \multicolumn{1}{r|}{139,071}       & \multicolumn{1}{r|}{16,808}      & 67,232                             \\ \cline{3-11} 
\multicolumn{1}{|c|}{}                                                              & \multicolumn{1}{c|}{}                               & \textbf{IC2}   & \multicolumn{1}{c|}{0.89}                  & \multicolumn{1}{c|}{0.74}               & 0.90                & 0.11                            & \multicolumn{1}{r|}{43,291}      & \multicolumn{1}{r|}{352,603}       & \multicolumn{1}{r|}{43,291}      & 173,164                            \\ \cline{3-11} 
\multicolumn{1}{|c|}{}                                                              & \multicolumn{1}{c|}{}                               & \textbf{CJ}    & \multicolumn{1}{c|}{0.90}                  & \multicolumn{1}{c|}{0.79}               & 0.90                & 0.06                            & \multicolumn{1}{r|}{124}         & \multicolumn{1}{r|}{2,008}         & \multicolumn{1}{r|}{124}         & 496                                \\ \cline{3-11} 
\multicolumn{1}{|c|}{}                                                              & \multicolumn{1}{c|}{}                               & \textbf{LGSR}  & \multicolumn{1}{c|}{0.94}                  & \multicolumn{1}{c|}{0.78}               & 0.94                & 0.01                            & \multicolumn{1}{r|}{3,338}       & \multicolumn{1}{r|}{282,042}       & \multicolumn{1}{r|}{3,338}       & 13,352                             \\ \cline{2-11} 
\multicolumn{1}{|c|}{}                                                              & \multicolumn{1}{c|}{\multirow{4}{*}{\textbf{B-SS}}} & \textbf{IC1*}  & \multicolumn{1}{c|}{0.92}                  & \multicolumn{1}{c|}{0.74}               & 0.94                & 0.11                            & \multicolumn{1}{r|}{16,808}      & \multicolumn{1}{r|}{139,071}       & \multicolumn{1}{r|}{540}         & 540                                \\ \cline{3-11} 
\multicolumn{1}{|c|}{}                                                              & \multicolumn{1}{c|}{}                               & \textbf{IC2*}  & \multicolumn{1}{c|}{0.89}                  & \multicolumn{1}{c|}{0.74}               & 0.90                & 0.11                            & \multicolumn{1}{r|}{43,291}      & \multicolumn{1}{r|}{352,603}       & \multicolumn{1}{r|}{509}         & 509                                \\ \cline{3-11} 
\multicolumn{1}{|c|}{}                                                              & \multicolumn{1}{c|}{}                               & \textbf{IC1**} & \multicolumn{1}{c|}{0.92}                  & \multicolumn{1}{c|}{0.74}               & 0.94                & 0.11                            & \multicolumn{1}{r|}{16,808}      & \multicolumn{1}{r|}{139,071}       & \multicolumn{1}{r|}{275}         & 275                                \\ \cline{3-11} 
\multicolumn{1}{|c|}{}                                                              & \multicolumn{1}{c|}{}                               & \textbf{IC2**} & \multicolumn{1}{c|}{0.89}                  & \multicolumn{1}{c|}{0.74}               & 0.90                & 0.11                            & \multicolumn{1}{r|}{43,291}      & \multicolumn{1}{r|}{352,603}       & \multicolumn{1}{r|}{270}         & 270                                \\ \hline
\multicolumn{1}{|c|}{\multirow{12}{*}{\textbf{C}}}                                  & \multicolumn{1}{c|}{\multirow{4}{*}{\textbf{1:1}}}  & \textbf{IC1}   & \multicolumn{1}{c|}{0.91}                  & \multicolumn{1}{c|}{0.78}               & 0.94                & 0.14                            & \multicolumn{1}{r|}{21,858}      & \multicolumn{1}{r|}{134,021}       & \multicolumn{1}{r|}{21,858}      & 21,858                             \\ \cline{3-11} 
\multicolumn{1}{|c|}{}                                                              & \multicolumn{1}{c|}{}                               & \textbf{IC2}   & \multicolumn{1}{c|}{0.89}                  & \multicolumn{1}{c|}{0.76}               & 0.90                & 0.13                            & \multicolumn{1}{r|}{52,989}      & \multicolumn{1}{r|}{342,905}       & \multicolumn{1}{r|}{52,989}      & 52,989                             \\ \cline{3-11} 
\multicolumn{1}{|c|}{}                                                              & \multicolumn{1}{c|}{}                               & \textbf{CJ}    & \multicolumn{1}{c|}{0.89}                  & \multicolumn{1}{c|}{0.80}               & 0.90                & 0.07                            & \multicolumn{1}{r|}{158}         & \multicolumn{1}{r|}{1,974}         & \multicolumn{1}{r|}{158}         & 158                                \\ \cline{3-11} 
\multicolumn{1}{|c|}{}                                                              & \multicolumn{1}{c|}{}                               & \textbf{LGSR}  & \multicolumn{1}{c|}{0.93}                  & \multicolumn{1}{c|}{0.79}               & 0.94                & 0.02                            & \multicolumn{1}{r|}{6,537}       & \multicolumn{1}{r|}{278,843}       & \multicolumn{1}{r|}{6,537}       & 6,537                              \\ \cline{2-11} 
\multicolumn{1}{|c|}{}                                                              & \multicolumn{1}{c|}{\multirow{4}{*}{\textbf{1:4}}}  & \textbf{IC1}   & \multicolumn{1}{c|}{0.91}                  & \multicolumn{1}{c|}{0.78}               & 0.94                & 0.14                            & \multicolumn{1}{r|}{21,858}      & \multicolumn{1}{r|}{134,021}       & \multicolumn{1}{r|}{21,858}      & 87,432                             \\ \cline{3-11} 
\multicolumn{1}{|c|}{}                                                              & \multicolumn{1}{c|}{}                               & \textbf{IC2}   & \multicolumn{1}{c|}{0.89}                  & \multicolumn{1}{c|}{0.76}               & 0.90                & 0.13                            & \multicolumn{1}{r|}{52,989}      & \multicolumn{1}{r|}{342,905}       & \multicolumn{1}{r|}{52,989}      & 211,956                            \\ \cline{3-11} 
\multicolumn{1}{|c|}{}                                                              & \multicolumn{1}{c|}{}                               & \textbf{CJ}    & \multicolumn{1}{c|}{0.89}                  & \multicolumn{1}{c|}{0.80}               & 0.90                & 0.07                            & \multicolumn{1}{r|}{158}         & \multicolumn{1}{r|}{1,974}         & \multicolumn{1}{r|}{158}         & 632                                \\ \cline{3-11} 
\multicolumn{1}{|c|}{}                                                              & \multicolumn{1}{c|}{}                               & \textbf{LGSR}  & \multicolumn{1}{c|}{0.93}                  & \multicolumn{1}{c|}{0.79}               & 0.94                & 0.02                            & \multicolumn{1}{r|}{6,537}       & \multicolumn{1}{r|}{278,843}       & \multicolumn{1}{r|}{6,537}       & 26,148                             \\ \cline{2-11} 
\multicolumn{1}{|c|}{}                                                              & \multicolumn{1}{c|}{\multirow{4}{*}{\textbf{B-SS}}} & \textbf{IC1*}  & \multicolumn{1}{c|}{0.91}                  & \multicolumn{1}{c|}{0.78}               & 0.94                & 0.14                            & \multicolumn{1}{r|}{21,858}      & \multicolumn{1}{r|}{134,021}       & \multicolumn{1}{r|}{690}         & 690                                \\ \cline{3-11} 
\multicolumn{1}{|c|}{}                                                              & \multicolumn{1}{c|}{}                               & \textbf{IC2*}  & \multicolumn{1}{c|}{0.89}                  & \multicolumn{1}{c|}{0.76}               & 0.90                & 0.13                            & \multicolumn{1}{r|}{52,989}      & \multicolumn{1}{r|}{342,905}       & \multicolumn{1}{r|}{636}         & 636                                \\ \cline{3-11} 
\multicolumn{1}{|c|}{}                                                              & \multicolumn{1}{c|}{}                               & \textbf{IC1**} & \multicolumn{1}{c|}{0.91}                  & \multicolumn{1}{c|}{0.78}               & 0.94                & 0.14                            & \multicolumn{1}{r|}{21,858}      & \multicolumn{1}{r|}{134,021}       & \multicolumn{1}{r|}{356}         & 356                                \\ \cline{3-11} 
\multicolumn{1}{|c|}{}                                                              & \multicolumn{1}{c|}{}                               & \textbf{IC2**} & \multicolumn{1}{c|}{0.89}                  & \multicolumn{1}{c|}{0.76}               & 0.90                & 0.13                            & \multicolumn{1}{r|}{52,989}      & \multicolumn{1}{r|}{342,905}       & \multicolumn{1}{r|}{326}         & 326                                \\ \hline
\end{tabular}%
}
\end{table}

\noindent
\begin{minipage}[h]{\textwidth}
\centering
\captionof{table}{Dataset statistics.}
\subfloat[Before filtering.]{
    \label{tab:dataset_statistics_before_filtering}
    \resizebox{0.48\linewidth}{!}{%
    \begin{tabular}{|r|r|r|r|c|c|}
    \hline
    \multicolumn{1}{|c|}{\textbf{\begin{tabular}[c]{@{}c@{}}Data\\ Name\end{tabular}}} & \multicolumn{1}{c|}{\textbf{\begin{tabular}[c]{@{}c@{}}Total\\ Weeks\end{tabular}}} & \multicolumn{1}{c|}{\textbf{\begin{tabular}[c]{@{}c@{}}Transactional\\ Weeks\end{tabular}}} & \multicolumn{1}{c|}{\textbf{\begin{tabular}[c]{@{}c@{}}Number\\ of\\ Customers\end{tabular}}} & \textbf{\begin{tabular}[c]{@{}c@{}}Average\\ Transactional\\ Weeks\\ per Customer\end{tabular}} & \textbf{\begin{tabular}[c]{@{}c@{}}Normalized\\ Average\\ Transactional\\ Weeks\\ per Customer\end{tabular}} \\ \hline
    IC1                                                                                & 78                                                                                  & 8,484,586                                                                                   & 490,138                                                                                       & 17.31                                                                                           & 0.22                                                                                                         \\ \hline
    IC2                                                                                & 115                                                                                 & 25,097,111                                                                                  & 937,507                                                                                       & 26.77                                                                                           & 0.23                                                                                                         \\ \hline
    CJ                                                                                 & 102                                                                                 & 123,888                                                                                     & 2,500                                                                                         & 49.56                                                                                           & 0.49                                                                                                         \\ \hline
    LGSR                                                                               & 117                                                                                 & 20,527,165                                                                                  & 494,880                                                                                       & 41.48                                                                                           & 0.35                                                                                                         \\ \hline
    \end{tabular}
    }
}
\hfill
\subfloat[After filtering.]{
    \label{tab:dataset_statistics_after_filtering}
    \resizebox{0.48\linewidth}{!}{%
    \begin{tabular}{|r|r|r|r|c|c|}
    \hline
    \multicolumn{1}{|c|}{\textbf{\begin{tabular}[c]{@{}c@{}}Data\\ Name\end{tabular}}} & \multicolumn{1}{c|}{\textbf{\begin{tabular}[c]{@{}c@{}}Total\\ Weeks\end{tabular}}} & \multicolumn{1}{c|}{\textbf{\begin{tabular}[c]{@{}c@{}}Transactional\\ Weeks\end{tabular}}} & \multicolumn{1}{c|}{\textbf{\begin{tabular}[c]{@{}c@{}}Number\\ of\\ Customers\end{tabular}}} & \textbf{\begin{tabular}[c]{@{}c@{}}Average\\ Transactional\\ Weeks\\ per Customer\end{tabular}} & \textbf{\begin{tabular}[c]{@{}c@{}}Normalized\\ Average\\ Transactional\\ Weeks\\ per Customer\end{tabular}} \\ \hline
    IC1                                                                                & 78                                                                                  & 6,507,930                                                                                   & 155,879                                                                                       & 41.75                                                                                           & 0.54                                                                                                         \\ \hline
    IC2                                                                                & 115                                                                                 & 21,642,961                                                                                  & 395,894                                                                                       & 54.67                                                                                           & 0.48                                                                                                         \\ \hline
    CJ                                                                                 & 102                                                                                 & 119,642                                                                                     & 2,132                                                                                         & 56.12                                                                                           & 0.55                                                                                                         \\ \hline
    LGSR                                                                               & 117                                                                                 & 19,318,828                                                                                  & 285,380                                                                                       & 67.70                                                                                            & 0.58                                                                                                         \\ \hline
    \end{tabular}
    }
}
\end{minipage}

\subsection{Model Training and Evaluation}
In this step, we train multiple machine-learning models on the prepared datasets and evaluate their performance using the selected classification metrics. 
Section~\ref{sec:implementation} provides implementation details, covering model architectures, hyperparameters, the experimental setup, libraries, and computational environment.

\subsection*{Dataset Information}
In total, four dataset sources were used: two proprietary datasets provided by Infocube (denoted as IC1 and IC2), and two publicly available datasets, \textit{The Complete Journey} and \textit{Let’s Get Sort of Real} (denoted as CJ and LGSR, respectively) \footnote{\url{https://www.dunnhumby.com/source-files/}}. The main characteristics of these sources are summarized in Table~\ref{tab:dataset_statistics_before_filtering} for the unfiltered datasets and in Table~\ref{tab:dataset_statistics_after_filtering} after filtering was applied. In these tables, the \textit{Transactional Weeks} field for each dataset source is calculated by counting, for each customer, the weeks in which they made at least one transaction, and then summing these counts across all customers. \textit{Average Transactional Weeks per Customer} gives the average number of transactional weeks per customer. \textit{Normalized Average Transactional Weeks per Customer} scales this value by the total period, showing the average transaction frequency per week per customer.

From the four dataset sources, we constructed 36 distinct datasets by applying the three churn-labeling definitions introduced in Subsection~\ref{subsec:churn_labeling}, together with the three balancing and subsampling strategies described in Subsection~\ref{subsec:balance_and_subsample}. Table~\ref{tab:data_desc_2} summarizes the resulting dataset families and their characteristics: the \textit{Cat.} column indicates whether a dataset is subsampled or balanced/imbalanced; N-CH and CH denote non-churners and churners, respectively. The \textit{Normalized Tenure} represents each customer’s active span divided by the total number of weeks. The churn ratio, along with the number of customers both before and after the balancing step, is also reported.

\section{Implementation}
\label{sec:implementation}

This section outlines the implementation details, experimental environment, and model-specific hyperparameters used in the study. All experiments were conducted using Python 3.10.9, with core data processing handled using pandas 1.5.3 and numpy 1.24.4. 

Sktime 0.16.1 was used for handling and converting time series data structures. Scikit-learn 1.7.0 was employed for evaluating model performance using metrics such as \textit{accuracy}, \textit{precision}, \textit{recall}, \textit{F1-score}, and \textit{ROC-AUC}, and also for implementing models such as Support Vector Machines and Random Forests. The implementation details and hyperparameter configurations for each model are described below.

\subsection{UniTS}
The UniTS models were implemented using the official UniTS framework, publicly available on GitHub\footnote{\url{https://github.com/mims-harvard/UniTS}}. All experiments were conducted using PyTorch 2.0.0, with GPU acceleration to enhance training efficiency.

The UniTS framework supports a variety of training paradigms, including prompt learning, few-shot learning, zero-shot inference, and supervised learning. 
\begin{itemize}
    \item \textbf{Prompt learning} enables the model to adapt to new tasks using learned task-specific prompt tokens, without modifying the underlying model weights.
    \item \textbf{Few-shot learning} allows the model to generalize to new tasks using only a small number of labeled examples.
    \item \textbf{Zero-shot inference} leverages the pre-trained model's representations to perform tasks for which it has not been explicitly trained.
    \item \textbf{Supervised learning} refers to traditional fine-tuning using fully labeled datasets for a specific downstream task.
\end{itemize}

In the original UniTS paper, the supervised learning approach yielded the best overall performance. Consequently, we adopt this approach in our study to showcase the full potential of the model. Rather than training from randomly initialized weights, we leveraged transfer learning by initializing the model with the publicly available checkpoint \texttt{units\_x64\_supervised\_checkpoint.pth}\footnote{\url{https://github.com/mims-harvard/UniTS/releases/tag/ckpt}}, obtained from the official UniTS release repository.

It is worth noting that we did not perform extensive hyperparameter tuning for UniTS. This decision aligns with the framework's design philosophy, which emphasizes plug-and-play usability over complex configuration.

The key hyperparameters and training settings used in our experiments are summarized in Table~\ref{tab:units_params}.
As discussed in Section~\ref{subsec:threats_transformers}, more targeted tuning of UniTS-specific hyperparameters or the use of different transformer-based architectures may yield different outcomes and insights. Therefore, the performance results presented here for UniTS should be taken as reflective of the potential performance of transformer-based architectures in general.

\subsection{XGBoost}
Experiments  with XGBoost were conducted using the \textbf{XGBClassifier} interface from version \textbf{3.0.2}. The model was trained using the binary logistic objective for churn classification. Feature values were used in their raw form without normalization, as the model is robust to unscaled input. The hyperparameters used are summarized in Table~\ref{tab:xgboost_params}.

\subsection{Support Vector Machine}
Support Vector Machine models were implemented using the \textbf{SVC} class from the scikit-learn library, version \textbf{1.7.0}. Before training, the input features were standardized using \textbf{StandardScaler}, which centers each feature around a zero mean and unit variance. This scaling step is essential for distance-based models like SVM to function properly.

The classifier was trained using a radial basis function (RBF) kernel. Table~\ref{tab:svm_params} lists the hyperparameters used.

\subsection{Random Forest}

Random Forest models were implemented using the \textit{RandomForestClassifier} class from scikit-learn version 1.7.0. No feature scaling was applied prior to training, as tree-based models are invariant to the scale of input features.

Table~\ref{tab:rf_params} shows the configuration used for the Random Forest model.

\begin{table}[h]
\centering
\caption{Model Hyperparameters.}
\begin{minipage}[t]{0.45\textwidth}
\centering
\subfloat[UniTS.]{
    \label{tab:units_params}
    \resizebox{\linewidth}{!}{%
    \begin{tabular}{|l|l|l|}
    \hline
    \textbf{Parameter} & \textbf{Value} & \textbf{Description} \\
    \hline
    prompt\_num & 10 & Number of prompts used in the model \\
    patch\_len & 16 & Patch length for input sequences \\
    stride & 16 & Stride for patching \\
    e\_layers & 3 & Number of encoder layers \\
    d\_model & 64 & Embedding dimension \\
    learning\_rate & 1e-4 & Fine-tuning learning rate \\
    weight\_decay & 5e-6 & L2 regularization \\
    train\_epochs & 10 & Number of training epochs \\
    batch\_size & 32 & Mini-batch size \\
    acc\_it & 32 & Gradient accumulation steps \\
    clip\_grad & 100 & Gradient clipping threshold \\
    fix\_seed & 42 & For reproducibility \\
    is\_training & 1 & Enables training mode \\
    lradj & supervised & Learning rate scheduler type \\
    task\_type & classification & Only classification task used \\
    \hline
    \end{tabular}%
    }
}
\vspace{2mm}
\subfloat[Random Forest.]{
    \label{tab:rf_params}
    \resizebox{0.8\linewidth}{!}{%
    \begin{tabular}{|l|l|l|}
    \hline
    \textbf{Parameter} & \textbf{Value} & \textbf{Description} \\
    \hline
    n\_estimators & 100 & Number of trees in the ensemble \\
    random\_state & 42 & For reproducibility \\
    \hline
    \end{tabular}%
    }
}
\end{minipage}
\hfill
\begin{minipage}[t]{0.52\textwidth}
\centering
\subfloat[XGBoost.]{
    \label{tab:xgboost_params}
    \resizebox{\linewidth}{!}{%
    \begin{tabular}{|l|l|l|}
    \hline
    \textbf{Parameter} & \textbf{Value} & \textbf{Description} \\
    \hline
    objective & binary:logistic & For binary classification \\
    n\_estimators & 100 & Number of boosting rounds \\
    max\_depth & 4 & Limits tree complexity to reduce overfitting \\
    learning\_rate & 0.1 & Moderate rate for stable convergence \\
    subsample & 0.8 & Random subset of data per tree \\
    colsample\_bytree & 0.8 & Subset of features per tree \\
    random\_state & 42 & For reproducibility \\
    eval\_metric & logloss & Log-loss used during training \\
    \hline
    \end{tabular}%
    }
}
\vspace{2mm}
\subfloat[SVM.]{
    \label{tab:svm_params}
    \resizebox{\linewidth}{!}{%
    \begin{tabular}{|l|l|l|}
    \hline
    \textbf{Parameter} & \textbf{Value} & \textbf{Description} \\
    \hline
    kernel & rbf & Radial basis function kernel \\
    C & 1.0 & Regularization parameter \\
    gamma & scale & Kernel coefficient based on number of features \\
    probability & True & Enables probability estimates (used for ROC-AUC) \\
    random\_state & 42 & For reproducibility \\
    \hline
    \end{tabular}%
    }
}
\end{minipage}
\end{table}








All experiments were conducted on a system with an AMD Ryzen 9 7950X CPU, 64 GB DDR5 RAM, two NVIDIA RTX 4090 GPUs, a 2 TB NVME SSD, running Ubuntu LTS 22.04.

\section{Results}
\label{sec:results}

In this section, we evaluate the models using standard performance metrics: Accuracy, Precision, Recall, F1-score, and AUC-ROC. These metrics are particularly relevant for churn prediction, where datasets are often imbalanced. In such cases, AUC-ROC provides a more reliable basis for performance comparison across models. Moreover, both false positives and false negatives carry significant business implications: excessive false positives may lead to unnecessary spending on customer retention, while false negatives can result in missed opportunities to intervene with actual churners. Consequently, Precision and Recall serve as critical complementary measures, providing deeper insights into these trade-offs.

The evaluation is conducted on 36 datasets derived from four raw data sources, as detailed in Section~\ref{sec:methodology}. To ensure a robust comparison, the datasets were systematically constructed with controlled variations in size, class imbalance, and churn definitions (A, B, and C). The corresponding results are reported in Table~\ref{tab:result_a}.

We observe that the tree-based models, \textit{XGBoost} and \textit{Random Forest}, consistently outperform both the transformer-based \textit{UniTS} architecture and the more conventional \textit{Support Vector Machine} (SVM) under all the conditions mentioned above.

We first contrast the performance under standard conditions with sufficient dataset size, namely the datasets generated from the data sources \textit{IC1}, \textit{IC2}, and \textit{LGSR} under balanced (1:1) and limited unbalanced (1:4) conditions. The datasets generated from the \textit{CJ} data source and all datasets under the \textit{B-SS} categories are considered data-scarce, i.e., due to the limited number of training examples, model performance is, in some cases, severely limited. We discuss and contrast model performance under these conditions separately in Subsection~\ref{subsec:impact_of_class_imbalance}.

When applying \textit{Churn Definition-A} to balanced datasets (1:1), \textit{XGBoost} achieves the highest accuracy of $0.77$ on \textit{IC1}, closely followed by \textit{Random Forest} at $0.76$, while \textit{UniTS} and \textit{SVM} remain lower at $0.72$. The superiority of tree-based models holds across \textit{IC2} and \textit{LGSR} as well. Accuracy is reported here, as it serves as a sufficient metric to evaluate performance under balanced conditions.

When shifting to the imbalanced datasets (1:4), we contrast precision and recall metrics to more effectively highlight performance differences under this imbalance. On \textit{IC1}, the precision ranking is as follows: \textit{Random Forest} ($0.76$), \textit{XGBoost} ($0.75$), \textit{SVM} ($0.70$), and \textit{UniTS} ($0.69$). For recall, \textit{XGBoost} achieves the highest value of $0.47$, followed by \textit{Random Forest} at $0.43$, \textit{UniTS} at $0.35$, and \textit{SVM} at only $0.09$. Similar patterns are observed on \textit{IC2} and \textit{LGSR}, confirming the superior performance of tree-based methods also under class imbalance.

Comparable trends hold under the alternative churn definitions B and C as well.

Below, we further analyze how model performance varies with class imbalance, dataset size, and different churn definitions, providing detailed insights into model-specific behavior.

\subsection{Performance by Churn Definition}

Increasing the churn threshold (moving from \textit{Churn Definition~A} to \textit{Churn Definition~C}) generally improves performance across all models. For example, on the 1:1 balanced IC1 dataset, UniTS achieves an accuracy of $0.72$ under \textit{Churn Definition~A}, which rises to $0.74$ under \textit{Churn Definition~B}, and further improves to $0.75$ under the most lenient threshold, \textit{Churn Definition~C}.

This increased performance could be due to an increase in the number of dataset samples used for training.

Note that since we use sub-sampling to limit dataset imbalance, increasing the size of the minority class, which is the churn class in this instance, increases the total number of samples used for training. As can be seen in Table~\ref{tab:data_desc_2}, by moving to a more lenient definition, from \textit{Churn Definition~A} to \textit{Churn Definition~C}, the number of customers classified as churners increases. For example, for the \textit{IC2} data source, the churn rate increases from 8\% under \textit{Churn Definition~A} to 11\% under \textit{Churn Definition~B}, and to 14\% under \textit{Churn Definition~C}. This corresponds to an increase in the total number of customers in the dataset from 24{,}850 to 43{,}716 between \textit{Churn Definitions A} and \textit{Churn Definition-C}.

Another reason for the improved performance could be that churn prediction under a more lenient definition is an easier task, as the classes are more easily separable. For \textit{Churn Definition~A}, both the monetary value and frequency of a customer must drop by 70\% between the observation and labeling window. Customers exhibiting such an aggressive drop may represent involuntary churners, e.g., those who have churned due to relocation or unexpected life events, where the weeks leading up to the churn event might not show strong behavioral signals indicative of churn.
\begin{table}[t]
\centering
\caption{Model results.}
\begin{minipage}{0.32\textwidth}
\centering
\label{tab:result_a}
\resizebox{\linewidth}{!}{%
\begin{tabular}{|c|c|c|c|c|c|c|}
\hline
\textbf{Cat.}                   & \textbf{Data}                   & \textbf{Metrics} & \textbf{UniTS} & \textbf{XGB}  & \textbf{RF}   & \textbf{SVM}  \\ \hline
\multirow{20}{*}{\textbf{1:1}}  & \multirow{5}{*}{\textbf{IC1}}   & Accuracy         & 0.72           & \textbf{0.77} & 0.76          & 0.72          \\ \cline{3-7} 
                                &                                 & Precision        & 0.72           & \textbf{0.77} & 0.75          & 0.74          \\ \cline{3-7} 
                                &                                 & Recall           & 0.74           & \textbf{0.78} & \textbf{0.78} & 0.67          \\ \cline{3-7} 
                                &                                 & F1-score         & 0.73           & \textbf{0.77} & 0.76          & 0.70          \\ \cline{3-7} 
                                &                                 & ROC-AUC          & 0.80           & \textbf{0.87} & \textbf{0.87} & 0.77          \\ \cline{2-7} 
                                & \multirow{5}{*}{\textbf{IC2}}   & Accuracy         & 0.78           & \textbf{0.81} & \textbf{0.81} & 0.77          \\ \cline{3-7} 
                                &                                 & Precision        & 0.78           & 0.81          & 0.81          & \textbf{0.82} \\ \cline{3-7} 
                                &                                 & Recall           & 0.77           & \textbf{0.82} & 0.80          & 0.69          \\ \cline{3-7} 
                                &                                 & F1-score         & 0.77           & \textbf{0.82} & 0.81          & 0.75          \\ \cline{3-7} 
                                &                                 & ROC-AUC          & 0.87           & \textbf{0.91} & \textbf{0.91} & 0.83          \\ \cline{2-7} 
                                & \multirow{5}{*}{\textbf{CJ}}    & Accuracy         & 0.52           & \textbf{0.72} & \textbf{0.72} & 0.62          \\ \cline{3-7} 
                                &                                 & Precision        & 0.57           & \textbf{0.85} & \textbf{0.85} & 0.69          \\ \cline{3-7} 
                                &                                 & Recall           & \textbf{0.76}  & 0.65          & 0.65          & 0.65          \\ \cline{3-7} 
                                &                                 & F1-score         & 0.65           & \textbf{0.73} & \textbf{0.73} & 0.67          \\ \cline{3-7} 
                                &                                 & ROC-AUC          & 0.50           & \textbf{0.76} & \textbf{0.76} & 0.65          \\ \cline{2-7} 
                                & \multirow{5}{*}{\textbf{LGSR}}  & Accuracy         & 0.80           & \textbf{0.94} & 0.93          & 0.83          \\ \cline{3-7} 
                                &                                 & Precision        & 0.80           & \textbf{0.93} & \textbf{0.93} & 0.82          \\ \cline{3-7} 
                                &                                 & Recall           & 0.84           & \textbf{0.95} & \textbf{0.95} & 0.90          \\ \cline{3-7} 
                                &                                 & F1-score         & 0.82           & \textbf{0.94} & \textbf{0.94} & 0.86          \\ \cline{3-7} 
                                &                                 & ROC-AUC          & 0.88           & \textbf{0.98} & \textbf{0.98} & 0.92          \\ \hline
\multirow{20}{*}{\textbf{1:4}}  & \multirow{5}{*}{\textbf{IC1}}   & Accuracy         & 0.84           & \textbf{0.86} & \textbf{0.86} & 0.81          \\ \cline{3-7} 
                                &                                 & Precision        & 0.69           & 0.75          & \textbf{0.76} & 0.70          \\ \cline{3-7} 
                                &                                 & Recall           & 0.35           & \textbf{0.47} & 0.43          & 0.09          \\ \cline{3-7} 
                                &                                 & F1-score         & 0.46           & \textbf{0.57} & 0.55          & 0.17          \\ \cline{3-7} 
                                &                                 & ROC-AUC          & 0.81           & \textbf{0.86} & \textbf{0.86} & 0.78          \\ \cline{2-7} 
                                & \multirow{5}{*}{\textbf{IC2}}   & Accuracy         & 0.87           & \textbf{0.87} & \textbf{0.87} & 0.83          \\ \cline{3-7} 
                                &                                 & Precision        & 0.73           & 0.74          & \textbf{0.76} & 0.71          \\ \cline{3-7} 
                                &                                 & Recall           & 0.52           & \textbf{0.57} & 0.51          & 0.24          \\ \cline{3-7} 
                                &                                 & F1-score         & 0.60           & \textbf{0.64} & 0.61          & 0.35          \\ \cline{3-7} 
                                &                                 & ROC-AUC          & 0.88           & \textbf{0.90} & \textbf{0.90} & 0.81          \\ \cline{2-7} 
                                & \multirow{5}{*}{\textbf{CJ}}    & Accuracy         & 0.28           & 0.94          & \textbf{0.96} & 0.86          \\ \cline{3-7} 
                                &                                 & Precision        & 0.11           & 0.80          & \textbf{0.89} & 0.00          \\ \cline{3-7} 
                                &                                 & Recall           & 0.60           & \textbf{0.80} & \textbf{0.80} & 0.00          \\ \cline{3-7} 
                                &                                 & F1-score         & 0.19           & 0.80          & \textbf{0.84} & 0.00          \\ \cline{3-7} 
                                &                                 & ROC-AUC          & 0.45           & 0.91          & \textbf{0.93} & 0.86          \\ \cline{2-7} 
                                & \multirow{5}{*}{\textbf{LGSR}}  & Accuracy         & 0.85           & \textbf{0.94} & \textbf{0.94} & 0.85          \\ \cline{3-7} 
                                &                                 & Precision        & 0.68           & 0.86          & \textbf{0.88} & 0.83          \\ \cline{3-7} 
                                &                                 & Recall           & 0.52           & \textbf{0.86} & 0.81          & 0.39          \\ \cline{3-7} 
                                &                                 & F1-score         & 0.59           & \textbf{0.86} & 0.85          & 0.53          \\ \cline{3-7} 
                                &                                 & ROC-AUC          & 0.86           & 0.97          & \textbf{0.98} & 0.90          \\ \hline
\multirow{20}{*}{\textbf{B-SS}} & \multirow{5}{*}{\textbf{IC1*}}  & Accuracy         & 0.49           & \textbf{0.73} & 0.72          & 0.67          \\ \cline{3-7} 
                                &                                 & Precision        & 0.52           & 0.75          & \textbf{0.76} & 0.75          \\ \cline{3-7} 
                                &                                 & Recall           & 0.67           & \textbf{0.75} & 0.72          & 0.59          \\ \cline{3-7} 
                                &                                 & F1-score         & 0.59           & \textbf{0.75} & 0.74          & 0.66          \\ \cline{3-7} 
                                &                                 & ROC-AUC          & 0.47           & \textbf{0.84} & \textbf{0.84} & 0.72          \\ \cline{2-7} 
                                & \multirow{5}{*}{\textbf{IC2*}}  & Accuracy         & 0.47           & \textbf{0.78} & \textbf{0.78} & 0.68          \\ \cline{3-7} 
                                &                                 & Precision        & 0.51           & \textbf{0.78} & \textbf{0.78} & 0.70          \\ \cline{3-7} 
                                &                                 & Recall           & 0.64           & \textbf{0.83} & \textbf{0.83} & 0.69          \\ \cline{3-7} 
                                &                                 & F1-score         & 0.57           & \textbf{0.81} & \textbf{0.81} & 0.70          \\ \cline{3-7} 
                                &                                 & ROC-AUC          & 0.48           & \textbf{0.87} & \textbf{0.87} & 0.74          \\ \cline{2-7} 
                                & \multirow{5}{*}{\textbf{IC1**}} & Accuracy         & 0.49           & \textbf{0.80} & 0.78          & 0.57          \\ \cline{3-7} 
                                &                                 & Precision        & 0.51           & \textbf{0.77} & 0.76          & 0.62          \\ \cline{3-7} 
                                &                                 & Recall           & 0.74           & \textbf{0.88} & 0.83          & 0.50          \\ \cline{3-7} 
                                &                                 & F1-score         & 0.60           & \textbf{0.82} & 0.80          & 0.55          \\ \cline{3-7} 
                                &                                 & ROC-AUC          & 0.50           & 0.85          & \textbf{0.86} & 0.69          \\ \cline{2-7} 
                                & \multirow{5}{*}{\textbf{IC2**}} & Accuracy         & 0.49           & 0.75          & \textbf{0.77} & 0.73          \\ \cline{3-7} 
                                &                                 & Precision        & 0.54           & 0.80          & \textbf{0.86} & 0.82          \\ \cline{3-7} 
                                &                                 & Recall           & \textbf{0.76}  & 0.74          & 0.72          & 0.66          \\ \cline{3-7} 
                                &                                 & F1-score         & 0.63           & 0.77          & \textbf{0.78} & 0.73          \\ \cline{3-7} 
                                &                                 & ROC-AUC          & 0.46           & 0.82          & \textbf{0.85} & 0.72          \\ \hline
\end{tabular}%
}
\par\vspace{6pt}
\footnotesize  (a) A (\(\alpha\) = 0.3, \(\beta\) = 0.3)
\end{minipage}
\hfill
\begin{minipage}{0.32\textwidth}
\centering
\label{tab:result_b}
\resizebox{\linewidth}{!}{%
\begin{tabular}{|c|c|c|c|c|c|c|}
\hline
\textbf{Cat.}                   & \textbf{Data}                   & \textbf{Metrics} & \textbf{UniTS} & \textbf{XGB}  & \textbf{RF}   & \textbf{SVM} \\ \hline
\multirow{20}{*}{\textbf{1:1}}  & \multirow{5}{*}{\textbf{IC1}}   & Accuracy         & 0.74           & \textbf{0.79} & 0.78          & 0.73         \\ \cline{3-7} 
                                &                                 & Precision        & 0.74           & 0.80          & \textbf{0.81} & 0.76         \\ \cline{3-7} 
                                &                                 & Recall           & 0.74           & \textbf{0.77} & 0.75          & 0.68         \\ \cline{3-7} 
                                &                                 & F1-score         & 0.74           & \textbf{0.78} & \textbf{0.78} & 0.72         \\ \cline{3-7} 
                                &                                 & ROC-AUC          & 0.82           & \textbf{0.89} & 0.88          & 0.79         \\ \cline{2-7} 
                                & \multirow{5}{*}{\textbf{IC2}}   & Accuracy         & 0.80           & \textbf{0.83} & 0.82          & 0.74         \\ \cline{3-7} 
                                &                                 & Precision        & 0.81           & \textbf{0.82} & \textbf{0.82} & 0.76         \\ \cline{3-7} 
                                &                                 & Recall           & 0.78           & \textbf{0.83} & 0.82          & 0.71         \\ \cline{3-7} 
                                &                                 & F1-score         & 0.79           & \textbf{0.83} & 0.82          & 0.73         \\ \cline{3-7} 
                                &                                 & ROC-AUC          & 0.88           & \textbf{0.91} & \textbf{0.91} & 0.82         \\ \cline{2-7} 
                                & \multirow{5}{*}{\textbf{CJ}}    & Accuracy         & 0.54           & \textbf{0.92} & \textbf{0.92} & 0.70         \\ \cline{3-7} 
                                &                                 & Precision        & 0.49           & \textbf{0.91} & 0.88          & 0.65         \\ \cline{3-7} 
                                &                                 & Recall           & 0.77           & 0.91          & \textbf{0.95} & 0.68         \\ \cline{3-7} 
                                &                                 & F1-score         & 0.60           & \textbf{0.91} & \textbf{0.91} & 0.67         \\ \cline{3-7} 
                                &                                 & ROC-AUC          & 0.62           & 0.96          & \textbf{0.97} & 0.77         \\ \cline{2-7} 
                                & \multirow{5}{*}{\textbf{LGSR}}  & Accuracy         & 0.80           & \textbf{0.95} & 0.94          & 0.84         \\ \cline{3-7} 
                                &                                 & Precision        & 0.80           & \textbf{0.92} & \textbf{0.92} & 0.79         \\ \cline{3-7} 
                                &                                 & Recall           & 0.79           & \textbf{0.97} & \textbf{0.97} & 0.91         \\ \cline{3-7} 
                                &                                 & F1-score         & 0.79           & \textbf{0.95} & \textbf{0.95} & 0.85         \\ \cline{3-7} 
                                &                                 & ROC-AUC          & 0.88           & \textbf{0.98} & \textbf{0.98} & 0.92         \\ \hline
\multirow{20}{*}{\textbf{1:4}}  & \multirow{5}{*}{\textbf{IC1}}   & Accuracy         & 0.85           & \textbf{0.87} & \textbf{0.87} & 0.83         \\ \cline{3-7} 
                                &                                 & Precision        & 0.71           & 0.76          & \textbf{0.77} & 0.75         \\ \cline{3-7} 
                                &                                 & Recall           & 0.46           & \textbf{0.53} & 0.49          & 0.22         \\ \cline{3-7} 
                                &                                 & F1-score         & 0.56           & \textbf{0.62} & 0.60          & 0.34         \\ \cline{3-7} 
                                &                                 & ROC-AUC          & 0.84           & \textbf{0.88} & \textbf{0.88} & 0.81         \\ \cline{2-7} 
                                & \multirow{5}{*}{\textbf{IC2}}   & Accuracy         & 0.87           & 0.88          & \textbf{0.87} & 0.83         \\ \cline{3-7} 
                                &                                 & Precision        & 0.73           & 0.73          & \textbf{0.75} & 0.71         \\ \cline{3-7} 
                                &                                 & Recall           & 0.56           & \textbf{0.61} & 0.55          & 0.23         \\ \cline{3-7} 
                                &                                 & F1-score         & 0.64           & \textbf{0.66} & 0.64          & 0.35         \\ \cline{3-7} 
                                &                                 & ROC-AUC          & 0.89           & \textbf{0.91} & \textbf{0.91} & 0.83         \\ \cline{2-7} 
                                & \multirow{5}{*}{\textbf{CJ}}    & Accuracy         & 0.35           & \textbf{0.93} & 0.92          & 0.85         \\ \cline{3-7} 
                                &                                 & Precision        & 0.17           & \textbf{0.93} & 1.00          & 1.00         \\ \cline{3-7} 
                                &                                 & Recall           & \textbf{0.71}  & 0.62          & 0.52          & 0.14         \\ \cline{3-7} 
                                &                                 & F1-score         & 0.27           & \textbf{0.74} & 0.69          & 0.25         \\ \cline{3-7} 
                                &                                 & ROC-AUC          & 0.51           & \textbf{0.98} & 0.97          & 0.85         \\ \cline{2-7} 
                                & \multirow{5}{*}{\textbf{LGSR}}  & Accuracy         & 0.87           & \textbf{0.94} & \textbf{0.94} & 0.86         \\ \cline{3-7} 
                                &                                 & Precision        & 0.71           & 0.84          & \textbf{0.85} & 0.81         \\ \cline{3-7} 
                                &                                 & Recall           & 0.59           & \textbf{0.87} & 0.84          & 0.42         \\ \cline{3-7} 
                                &                                 & F1-score         & 0.65           & \textbf{0.86} & 0.85          & 0.56         \\ \cline{3-7} 
                                &                                 & ROC-AUC          & 0.90           & \textbf{0.98} & \textbf{0.98} & 0.92         \\ \hline
\multirow{20}{*}{\textbf{B-SS}} & \multirow{5}{*}{\textbf{IC1*}}  & Accuracy         & 0.47           & \textbf{0.77} & 0.75          & 0.69         \\ \cline{3-7} 
                                &                                 & Precision        & 0.46           & \textbf{0.73} & \textbf{0.73} & 0.65         \\ \cline{3-7} 
                                &                                 & Recall           & 0.73           & \textbf{0.80} & 0.76          & 0.72         \\ \cline{3-7} 
                                &                                 & F1-score         & 0.56           & \textbf{0.76} & 0.74          & 0.69         \\ \cline{3-7} 
                                &                                 & ROC-AUC          & 0.55           & \textbf{0.86} & \textbf{0.86} & 0.75         \\ \cline{2-7} 
                                & \multirow{5}{*}{\textbf{IC2*}}  & Accuracy         & 0.49           & 0.77          & \textbf{0.78} & 0.73         \\ \cline{3-7} 
                                &                                 & Precision        & 0.50           & 0.75          & \textbf{0.76} & 0.75         \\ \cline{3-7} 
                                &                                 & Recall           & 0.74           & 0.82          & \textbf{0.83} & 0.70         \\ \cline{3-7} 
                                &                                 & F1-score         & 0.60           & \textbf{0.79} & \textbf{0.79} & 0.73         \\ \cline{3-7} 
                                &                                 & ROC-AUC          & 0.49           & 0.86          & \textbf{0.87} & 0.77         \\ \cline{2-7} 
                                & \multirow{5}{*}{\textbf{IC1**}} & Accuracy         & 0.55           & 0.78          & \textbf{0.85} & 0.76         \\ \cline{3-7} 
                                &                                 & Precision        & 0.58           & 0.82          & \textbf{0.87} & 0.80         \\ \cline{3-7} 
                                &                                 & Recall           & 0.78           & 0.79          & \textbf{0.87} & 0.78         \\ \cline{3-7} 
                                &                                 & F1-score         & 0.66           & 0.81          & \textbf{0.87} & 0.79         \\ \cline{3-7} 
                                &                                 & ROC-AUC          & 0.50           & 0.88          & \textbf{0.90} & 0.83         \\ \cline{2-7} 
                                & \multirow{5}{*}{\textbf{IC2**}} & Accuracy         & 0.54           & \textbf{0.81} & \textbf{0.81} & 0.66         \\ \cline{3-7} 
                                &                                 & Precision        & 0.57           & 0.84          & \textbf{0.88} & 0.71         \\ \cline{3-7} 
                                &                                 & Recall           & 0.72           & \textbf{0.84} & 0.75          & 0.66         \\ \cline{3-7} 
                                &                                 & F1-score         & 0.64           & \textbf{0.84} & 0.81          & 0.68         \\ \cline{3-7} 
                                &                                 & ROC-AUC          & 0.61           & 0.91          & \textbf{0.91} & 0.78         \\ \hline
\end{tabular}%
}
\par\vspace{6pt}
\footnotesize  (b) B (\(\alpha\) = 0.4, \(\beta\) = 0.45)
\end{minipage}
\hfill
\begin{minipage}{0.32\textwidth}
\centering
\label{tab:result_c}
\resizebox{\linewidth}{!}{%
\begin{tabular}{|c|c|c|c|c|c|c|}
\hline
\textbf{Cat.}                   & \textbf{Data}                   & \textbf{Metrics} & \textbf{UniTS} & \textbf{XGB}  & \textbf{RF}   & \textbf{SVM} \\ \hline
\multirow{20}{*}{\textbf{1:1}}  & \multirow{5}{*}{\textbf{IC1}}   & Accuracy         & 0.75           & \textbf{0.80} & \textbf{0.80} & 0.75         \\ \cline{3-7} 
                                &                                 & Precision        & 0.74           & \textbf{0.81} & \textbf{0.81} & 0.77         \\ \cline{3-7} 
                                &                                 & Recall           & 0.77           & \textbf{0.79} & 0.78          & 0.71         \\ \cline{3-7} 
                                &                                 & F1-score         & 0.76           & \textbf{0.80} & \textbf{0.80} & 0.74         \\ \cline{3-7} 
                                &                                 & ROC-AUC          & 0.82           & \textbf{0.89} & \textbf{0.89} & 0.81         \\ \cline{2-7} 
                                & \multirow{5}{*}{\textbf{IC2}}   & Accuracy         & 0.80           & \textbf{0.84} & 0.83          & 0.79         \\ \cline{3-7} 
                                &                                 & Precision        & 0.80           & \textbf{0.83} & \textbf{0.83} & 0.83         \\ \cline{3-7} 
                                &                                 & Recall           & 0.80           & \textbf{0.85} & 0.83          & 0.72         \\ \cline{3-7} 
                                &                                 & F1-score         & 0.80           & \textbf{0.84} & 0.83          & 0.77         \\ \cline{3-7} 
                                &                                 & ROC-AUC          & 0.89           & \textbf{0.92} & \textbf{0.92} & 0.85         \\ \cline{2-7} 
                                & \multirow{5}{*}{\textbf{CJ}}    & Accuracy         & 0.56           & \textbf{0.81} & \textbf{0.81} & 0.75         \\ \cline{3-7} 
                                &                                 & Precision        & 0.54           & \textbf{0.81} & \textbf{0.81} & 0.70         \\ \cline{3-7} 
                                &                                 & Recall           & \textbf{0.84}  & 0.81          & 0.81          & 0.88         \\ \cline{3-7} 
                                &                                 & F1-score         & 0.66           & \textbf{0.81} & \textbf{0.81} & 0.78         \\ \cline{3-7} 
                                &                                 & ROC-AUC          & 0.61           & 0.85          & \textbf{0.86} & 0.79         \\ \cline{2-7} 
                                & \multirow{5}{*}{\textbf{LGSR}}  & Accuracy         & 0.79           & \textbf{0.93} & \textbf{0.93} & 0.83         \\ \cline{3-7} 
                                &                                 & Precision        & 0.76           & \textbf{0.91} & \textbf{0.91} & 0.78         \\ \cline{3-7} 
                                &                                 & Recall           & 0.87           & \textbf{0.96} & \textbf{0.96} & 0.92         \\ \cline{3-7} 
                                &                                 & F1-score         & 0.81           & \textbf{0.94} & 0.93          & 0.84         \\ \cline{3-7} 
                                &                                 & ROC-AUC          & 0.89           & \textbf{0.97} & \textbf{0.97} & 0.91         \\ \hline
\multirow{20}{*}{\textbf{1:4}}  & \multirow{5}{*}{\textbf{IC1}}   & Accuracy         & 0.86           & \textbf{0.88} & 0.87          & 0.84         \\ \cline{3-7} 
                                &                                 & Precision        & 0.70           & 0.74          & \textbf{0.76} & 0.73         \\ \cline{3-7} 
                                &                                 & Recall           & 0.49           & \textbf{0.57} & 0.53          & 0.27         \\ \cline{3-7} 
                                &                                 & F1-score         & 0.58           & \textbf{0.65} & 0.63          & 0.40         \\ \cline{3-7} 
                                &                                 & ROC-AUC          & 0.86           & 0.89          & \textbf{0.90} & 0.82         \\ \cline{2-7} 
                                & \multirow{5}{*}{\textbf{IC2}}   & Accuracy         & 0.87           & \textbf{0.88} & \textbf{0.88} & 0.83         \\ \cline{3-7} 
                                &                                 & Precision        & 0.72           & 0.74          & \textbf{0.75} & 0.70         \\ \cline{3-7} 
                                &                                 & Recall           & 0.59           & \textbf{0.64} & 0.59          & 0.30         \\ \cline{3-7} 
                                &                                 & F1-score         & 0.65           & \textbf{0.69} & 0.66          & 0.42         \\ \cline{3-7} 
                                &                                 & ROC-AUC          & 0.90           & \textbf{0.92} & \textbf{0.92} & 0.84         \\ \cline{2-7} 
                                & \multirow{5}{*}{\textbf{CJ}}    & Accuracy         & 0.39           & \textbf{0.90} & 0.85          & 0.80         \\ \cline{3-7} 
                                &                                 & Precision        & 0.24           & \textbf{0.88} & 0.83          & 0.80         \\ \cline{3-7} 
                                &                                 & Recall           & \textbf{0.80}  & 0.63          & 0.43          & 0.11         \\ \cline{3-7} 
                                &                                 & F1-score         & 0.37           & \textbf{0.73} & 0.57          & 0.20         \\ \cline{3-7} 
                                &                                 & ROC-AUC          & 0.61           & \textbf{0.94} & \textbf{0.94} & 0.83         \\ \cline{2-7} 
                                & \multirow{5}{*}{\textbf{LGSR}}  & Accuracy         & 0.87           & \textbf{0.94} & 0.93          & 0.86         \\ \cline{3-7} 
                                &                                 & Precision        & 0.71           & 0.82          & \textbf{0.83} & 0.76         \\ \cline{3-7} 
                                &                                 & Recall           & 0.54           & \textbf{0.87} & 0.82          & 0.43         \\ \cline{3-7} 
                                &                                 & F1-score         & 0.61           & \textbf{0.84} & 0.83          & 0.55         \\ \cline{3-7} 
                                &                                 & ROC-AUC          & 0.90           & \textbf{0.97} & \textbf{0.97} & 0.92         \\ \hline
\multirow{20}{*}{\textbf{B-SS}} & \multirow{5}{*}{\textbf{IC1*}}  & Accuracy         & 0.73           & 0.76          & \textbf{0.78} & 0.71         \\ \cline{3-7} 
                                &                                 & Precision        & 0.79           & 0.77          & \textbf{0.80} & 0.71         \\ \cline{3-7} 
                                &                                 & Recall           & 0.63           & \textbf{0.76} & \textbf{0.76} & 0.74         \\ \cline{3-7} 
                                &                                 & F1-score         & 0.70           & 0.76          & \textbf{0.78} & 0.72         \\ \cline{3-7} 
                                &                                 & ROC-AUC          & 0.79           & \textbf{0.88} & \textbf{0.88} & 0.78         \\ \cline{2-7} 
                                & \multirow{5}{*}{\textbf{IC2*}}  & Accuracy         & 0.73           & \textbf{0.85} & \textbf{0.85} & 0.75         \\ \cline{3-7} 
                                &                                 & Precision        & 0.78           & \textbf{0.87} & 0.86          & 0.73         \\ \cline{3-7} 
                                &                                 & Recall           & 0.64           & \textbf{0.84} & 0.83          & 0.81         \\ \cline{3-7} 
                                &                                 & F1-score         & 0.70           & \textbf{0.85} & \textbf{0.85} & 0.77         \\ \cline{3-7} 
                                &                                 & ROC-AUC          & 0.80           & \textbf{0.93} & \textbf{0.93} & 0.84         \\ \cline{2-7} 
                                & \multirow{5}{*}{\textbf{IC1**}} & Accuracy         & 0.47           & \textbf{0.75} & 0.71          & 0.66         \\ \cline{3-7} 
                                &                                 & Precision        & 0.47           & \textbf{0.77} & 0.72          & 0.63         \\ \cline{3-7} 
                                &                                 & Recall           & \textbf{0.72}  & 0.68          & 0.67          & 0.71         \\ \cline{3-7} 
                                &                                 & F1-score         & 0.57           & \textbf{0.72} & 0.69          & 0.67         \\ \cline{3-7} 
                                &                                 & ROC-AUC          & 0.49           & 0.82          & \textbf{0.83} & 0.71         \\ \cline{2-7} 
                                & \multirow{5}{*}{\textbf{IC2**}} & Accuracy         & 0.50           & 0.82          & \textbf{0.83} & 0.69         \\ \cline{3-7} 
                                &                                 & Precision        & 0.53           & 0.83          & \textbf{0.84} & 0.72         \\ \cline{3-7} 
                                &                                 & Recall           & 0.68           & 0.83          & \textbf{0.86} & 0.74         \\ \cline{3-7} 
                                &                                 & F1-score         & 0.60           & 0.83          & \textbf{0.85} & 0.73         \\ \cline{3-7} 
                                &                                 & ROC-AUC          & 0.54           & 0.86          & \textbf{0.86} & 0.78         \\ \hline
\end{tabular}%
}
\par\vspace{6pt}
\footnotesize (c) C (\(\alpha\) = 0.5, \(\beta\) = 0.5)
\end{minipage}

\end{table}

\subsection{Impact of Class Imbalance} \label{subsec:impact_of_class_imbalance}

In imbalanced classification settings, accuracy becomes a misleading metric, as any apparent gain is largely driven by the dominance of the non-churner class. To properly capture the models’ ability to detect churners, we instead evaluate performance using precision and recall. Precision measures how many of the predicted churners are truly churners, while recall measures how many of the actual churners the model successfully identifies. These metrics are more informative under imbalance, as they directly reflect performance on the minority class, which is the primary focus in churn prediction for real-world scenarios.

When transitioning from balanced (1:1) to imbalanced (1:4) datasets, model performance generally decreases as the class ratio becomes more skewed. The observed trend shows that this decline is most pronounced for models that are highly sensitive to class distribution, such as SVM, while tree-based methods (XGB, RF) and the transformer-based UniTS demonstrate greater robustness.

For example, under the IC2 dataset with \textit{Churn Definition~A}, the following results are observed when moving from a balanced to an imbalanced setting. For UniTS, precision decreases from $0.78$ to $0.73$ and recall drops from $0.77$ to $0.52$. For XGB, precision declines from $0.81$ to $0.74$ and recall decreases from $0.82$ to $0.57$. For RF, precision goes from $0.81$ to $0.76$ and recall falls from $0.80$ to $0.51$. Finally, for SVM, precision decreases from $0.82$ to $0.71$, but recall collapses dramatically from $0.69$ to $0.24$.

These results reveal a clear pattern: while precision remains relatively stable across all models, recall suffers a substantial decline under imbalance. This indicates that models become less effective at identifying churners as the minority class becomes underrepresented. Among the evaluated methods, tree-based models (XGB and RF) and the transformer-based UniTS demonstrate better resilience, maintaining moderate recall values even when the class distribution is skewed.

SVM experiences the most severe performance degradation under imbalance because, in the soft-margin setting, the separating hyperplane becomes skewed toward the minority class. As imbalance grows, the ratio of positive to negative support vectors becomes more imbalanced and samples near the margin are more likely to be classified as negative. While its precision decreases moderately ($0.82 \to 0.71$), recall collapses to $0.24$ because the model fails to predict churners effectively. This consistent trend demonstrates SVM’s sensitivity to class imbalance and it's worse performance compared to XGB, RF, and UniTS, which handle skewed distributions more robustly.

\subsection{Small Dataset and Data Efficiency Analysis}

Since the datasets considered in this section are balanced (1:1), the accuracy metric is sufficient for comparing performance. As the dataset size decreases, the performance of all models generally declines, although the degree of degradation under data scarcity varies significantly across models.

As a representative example, we examine the IC2 dataset under \textit{Churn Definition~C} (see Table~\ref{tab:data_desc_2}), where IC2 represents the full dataset, IC2* is a considerably smaller subset, and IC2** is approximately half the size of IC2*. Under these settings, UniTS achieves an accuracy of $0.80$ on the full IC2 dataset but drops to $0.73$ under IC2*, and further collapses to $0.50$ on IC2**. XGB achieves $0.84$ accuracy on IC2 and maintains a strong $0.85$ on IC2*, before slightly decreasing to $0.82$ under IC2**. Similarly, RF obtains $0.83$ accuracy on IC2, remains stable at $0.85$ on IC2*, and retains $0.83$ on IC2**. In contrast, SVM performs reasonably well on the full IC2 dataset with $0.79$ accuracy and drops slightly to $0.75$ under IC2*, but declines further to $0.69$ on IC2**. While these numbers are reported for IC2 as an example, similar patterns are consistently observed across other datasets.

These results reveal a clear and consistent pattern: tree-based models (XGB and RF) exhibit strong robustness under limited data availability, retaining high accuracy even when datasets are significantly reduced. SVM shows moderate resilience but suffers more noticeable degradation in extremely low-resource scenarios. UniTS, on the other hand, demonstrates a consistent trend of severe performance deterioration when data becomes scarce, confirming that it struggles to generalize effectively without sufficient training samples.

Transformer-based models like UniTS require larger datasets due to their higher complexity and large parameter count. Unlike tree-based methods, which generalize well with limited samples, transformers need abundant data to effectively capture temporal patterns and learn stable representations. In scarce-resource scenarios, the limited number of examples reduces the diversity of behavioral signals, making it harder for the model to distinguish churners from non-churners.

\section{Threats to Validity}
\label{sec:threats}

This section discusses potential limitations and factors that may affect the validity and generalizability of the findings presented in this study.

\subsection*{Limitations of Dataset, Applied Churn Definition, and Filtering Techniques}

Although the experiments were conducted on multiple datasets and under different churn definitions, the scope of this study remains limited, and the results may not fully generalize to all churn prediction tasks in retail contexts. In domains where consumer behavior is more complex, the UniTS model could have greater opportunities to capture intricate interdependencies, potentially leading to improved performance.

Moreover, while several churn labeling strategies were explored based on Infocube's requirements, the study does not cover the full spectrum of possible churn definitions. Alternative labeling schemes may substantially change the framing of the classification task and, in turn, affect model performance.

Finally, the dataset filtering procedure also introduces limitations to the applicability of our results, as customers with fewer than 20 transactional weeks were excluded. While this threshold was justified in our case, ensuring that the analysis focused on customers with sufficient transactional history to produce more reliable predictions, it can restrict the transferability of findings to other domains. For instance, in e-commerce or other retail environments where shorter customer life-cycles are common and business objectives may differ, models could produce different outcomes when applied to customers with limited histories.

\subsection*{Considerations Regarding Transformer-Based Methods}
\label{subsec:threats_transformers}

This work does not attempt to exhaustively optimize transformer-based models for churn prediction. Since UniTS is designed as a user-friendly, plug-and-play platform without the need for extensive model tuning, the reported results are valid for performance comparisons under default configurations. However, it remains possible that transformer-based approaches with architecture, optimizers, and hyperparameters specifically fine-tuned for churn prediction tasks could achieve superior results.

\section{Conclusion}
\label{sec:conclusion}

This study presents a comprehensive comparison between conventional machine learning models and the transformer-based UniTS model for the task of churn prediction in non-contractual retail environments. By conducting a rigorous evaluation of the models across 36 datasets with varying levels of imbalance, size, and churn definitions, we shed light on the performance trade-offs between classical and modern AI techniques in real-world retail applications.

Our findings demonstrate that tree-based models, particularly XGBoost and Random Forest, consistently outperform UniTS and SVM in most experimental settings. These models offer high predictive accuracy, robustness to class imbalance, and remarkable data efficiency, achieving strong performance even with as few as 200–600 churners. In contrast, the UniTS model, while theoretically well-suited for modeling temporal dynamics, underperforms on small to medium-sized tabular datasets, likely due to its high data requirements.

From a practical perspective, the results suggest that traditional ensemble methods remain a reliable and efficient choice for churn prediction in typical retail environments where data is structured and often limited. Although transformer-based models may hold promise, particularly in scenarios involving large-scale behavioral data and rich temporal histories, their advantages are not fully realized under constrained data settings.

Looking ahead, future research could explore how UniTS and similar architectures perform when additional behavioral and contextual features (e.g., browsing activity, promotions, seasonality) are included. Additionally, a systematic analysis of training time, energy efficiency, and deployment constraints would help assess the suitability of different model types for modern AI-accelerated infrastructures. By advancing our understanding of the conditions under which advanced architectures can provide tangible benefits, we hope to guide more effective and scalable applications of AI in customer retention strategies.

\begin{credits}
\subsubsection{\ackname} This work was carried out while Syed Saad Saif was enrolled in the Italian National Doctorate on Artificial Intelligence run by Sapienza University of Rome in collaboration with UnitelmaSapienza University of Rome, with his doctoral fellowship co-funded by Infocube SRL.

\subsubsection{\discintname}
The authors declare that they have no competing interests. 
\end{credits}
%
%
%
\bibliographystyle{splncs04}
\bibliography{biblography}
\end{document}